%% file: main.tex
\documentclass[conference]{IEEEtran}

\usepackage{amsmath,amssymb,amsfonts}
\usepackage{graphicx}
\usepackage{textcomp}
\usepackage{xcolor}
\usepackage{tikz}
\usetikzlibrary{positioning, arrows.meta, calc, fit, backgrounds, shapes.geometric}
\usepackage{cite}
\usepackage{hyperref}
\usepackage{booktabs}
\usepackage{multirow}
\usepackage{algorithm}
\usepackage{algpseudocode}
\usepackage{subcaption}
\usepackage{bbm}          
\usepackage{balance}       
\usepackage{url}
\usepackage{tcolorbox}
\usepackage{listings}
\tcbuselibrary{listings,breakable}
\usepackage{comment}

\usepackage[T1]{fontenc}
\newcommand{\sys}{\textit{\textbf{CORAL}}}
\graphicspath{{figures/}}

\newcommand{\placeholder}[1]{\textcolor{red}{\textbf{[#1]}}}

\begin{document}

\title{\sys: \textbf{CO}ntextual \textbf{R}easoning \textbf{A}nd \textbf{L}ocal Planning in A Hierarchical VLM Framework for Underwater Monitoring}
\author{
\IEEEauthorblockN{
Zhenqi Wu\IEEEauthorrefmark{1}\IEEEauthorrefmark{3},
Yuanjie Lu\IEEEauthorrefmark{2},
Xuesu Xiao\IEEEauthorrefmark{2},
Xiaomin Lin\IEEEauthorrefmark{1}
}

\IEEEauthorblockA{\IEEEauthorrefmark{1}
University of South Florida}

\IEEEauthorblockA{\IEEEauthorrefmark{2}
George Mason University}

\IEEEauthorblockA{\IEEEauthorrefmark{3}
Corresponding Author\\
Email: zhenqi@usf.edu}
}

\maketitle

\begin{abstract}
Oyster reefs are critical ecosystem species that sustain biodiversity, filter water, and protect coastlines, yet they continue to decline globally. Restoring these ecosystems requires regular underwater monitoring to assess reef health, a task that remains costly, hazardous, and limited when performed by human divers. Autonomous underwater vehicles (AUVs) offer a promising alternative, but existing AUVs rely on geometry-based navigation that cannot interpret scene semantics. Recent vision-language models (VLMs) enable semantic reasoning for intelligent exploration, but existing VLM-driven systems adopt an end-to-end paradigm, introducing three key limitations. First, these systems require the VLM to generate every navigation decision, forcing frequent waits for inference. Second, VLMs cannot model robot dynamics, causing collisions in cluttered environments. Third, limited self-correction allows small deviations to accumulate into large path errors. To address these limitations, we propose \sys, a framework that decouples high-level semantic reasoning from low-level reactive control. The VLM provides high-level exploration guidance by selecting waypoints, while a dynamics-based planner handles low-level collision-free execution. A geometric verification module validates waypoints and triggers replanning when needed. 
Compared with the previous state-of-the-art, \sys~improves coverage by 14.28\% percentage points, or 17.85\% relatively, reduces collisions by 100\%, and requires 57\% fewer VLM calls.
\end{abstract}

\input{sections/introduction}
\input{sections/related_work}
\input{sections/approach}

\input{sections/experiments}

\input{sections/results}

\input{sections/conclusion}

\bibliographystyle{IEEEtran}
\bibliography{references}

\input{sections/appendix}

\end{document}

%% file: sections/introduction.tex
\section{Introduction}
\label{sec:introduction}

Oyster reefs are critical ecosystem species that sustain biodiversity, filter water, and protect coastlines, yet 85\% have been lost globally~\cite{beck2011oyster}, making reef restoration increasingly urgent~\cite{howie2021contemporary}. Restoration success requires repeated underwater surveys to assess recovery progress, but human-conducted monitoring remains costly, hazardous, and spatially limited due to dive time, personnel, and vessel constraints~\cite{cardenas2024robotic}. Autonomous underwater vehicles (AUVs) offer a solution, but navigating complex reef environments with limited sensing and localization remains a fundamental challenge~\cite{zhang2023auvreview}.
\begin{figure}[t]
    \centering
    {\includegraphics[width=\columnwidth]{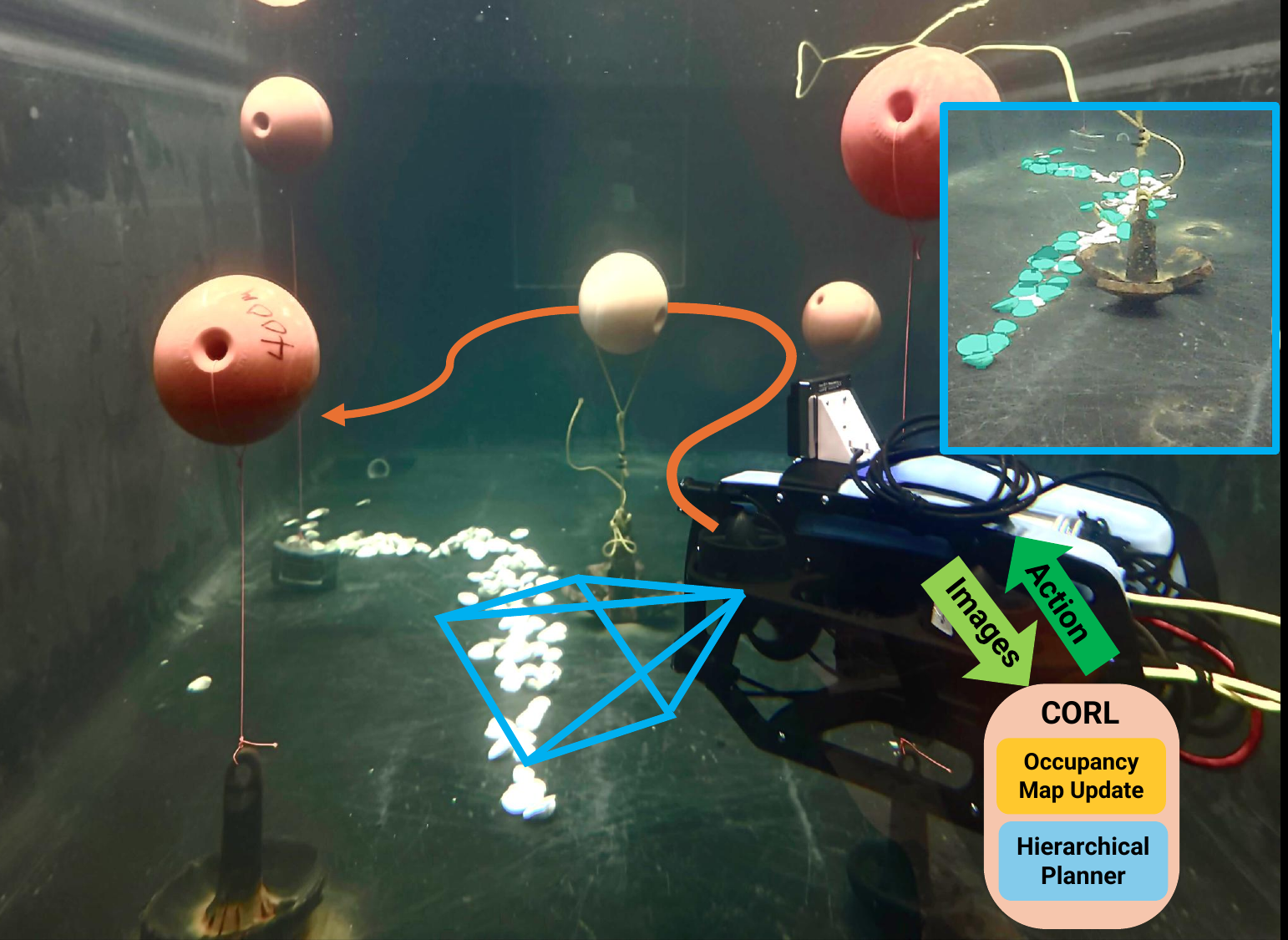}}
    \vspace{-3mm}
    \caption{Example of \sys~ deployed in the real world on a BlueROV surveying an oyster reef in a pool. }
    \label{fig:beauty_shot}
    \vspace{-4mm}
\end{figure}
Traditional AUV navigation methods, similar to mobile robot navigation, adopt geometry-based approaches that represent the environment as occupancy maps and execute pre-planned paths~\cite{xiao2022motion, christensen2022underwater}. While effective for collision-free navigation in open environments, these methods perceive the environment purely through geometric primitives and cannot distinguish ecologically meaningful structures (e.g., coral reefs) from generic terrain. Consequently, they treat all regions equally and cannot prioritize exploration toward ecologically valuable areas when locations are unknown a priori.

Learning-based approaches partially address these limitations by enabling adaptive, perception-driven navigation. For example, UIVNAV~\cite{lin2024uivnav} uses imitation learning to steer the robot toward objects of interest while avoiding obstacles. However, as reactive policies operating on local visual observations, such methods lack long-horizon planning and semantic reasoning capabilities needed to systematically explore complex reef topologies with branching paths, dead ends, or spatially distributed clusters.

Recent vision language models (VLMs) combine visual perception with commonsense reasoning and have shown promise for underwater autonomy. OceanPlan~\cite{yang2024oceanplan} enables natural language AUV piloting, while DREAM~\cite{wu2025dream} supports reef monitoring through VLM guided exploration with occupancy maps and Chain of Thought reasoning. However, most VLM based underwater navigation methods~\cite{wang2025underwatervla} rely on an end to end design in which the VLM makes every navigation decision, leading to three main limitations: low efficiency due to frequent inference delays, weak dynamics awareness that increases collision risk in constrained reef environments, and limited self correction that allows small errors to accumulate into large path deviations.

To address these limitations, we propose \sys, a hierarchical framework combining VLM semantic reasoning with dynamics-aware control. The VLM acts as a high-level exploration advisor, generating waypoints based on scene understanding, while a dynamics-based planner, Decremental Dynamics Planning (DDP)~\cite{lu2025ddp}, serves as the low-level controller, continuously executing collision-free trajectories. A path feasibility monitor tracks execution deviations and triggers replanning when thresholds are exceeded, closing the self-correction loop. Our main contributions are:

\begin{itemize}
    \item A hierarchical VLM-guided navigation framework for underwater vehicles that decouples semantic waypoint selection from dynamics-based trajectory execution, enabling asynchronous operation without per-step VLM queries.
    
    \item Integration of visual semantics with dynamics-based planning (DDP), where underwater vehicle state is propagated through nonlinear underactuated dynamics models with fine-grained collision checking against vision-derived obstacle maps at each integration step, synthesizing dynamically admissible trajectories that respect both underactuated constraints and thruster saturation limits.

    \item A path feasibility monitor that tracks deviations between planned and executed trajectories, providing structured feedback to trigger replanning when thresholds are exceeded, closing the self-correction loop.

    \item We evaluate \sys{} on simulated reef environments of various complexity and deploy in real-time in a pool demonstrating the feasibility of real-world application(Fig.~\ref{fig:beauty_shot}). 
    
\end{itemize}


%% file: sections/related_work.tex
\section{Related Work}
\label{sec:related}


\subsection{Classical Geometry-Based Navigation}
\label{sec:rw_classical}
Autonomous underwater monitoring requires a navigation system that is both stable and adaptive. Classical navigation methods for mobile robots~\cite{xiaoautonomous, lu2025adaptive, xu2025verti} construct geometric representations such as occupancy grids from sensor observations and plan collision-free paths in real time. 
These approaches have been widely adopted in underwater exploration, where AUVs combine occupancy-based mapping with reactive obstacle avoidance and execute pre-programmed coverage paths for tasks like ocean surveying and ecosystem monitoring~\cite{christensen2022underwater, yamauchi1997frontier, wynn2014auv_marine}. However, these geometry-driven planners reason only about free space and obstacles. They cannot identify or prioritize ecologically relevant targets such as reef clusters when locations are unknown a priori. This limits the effectiveness for adaptive, semantic-aware monitoring tasks.

\subsection{Learning-Based Semantic Navigation}
\label{sec:rw_learning}

Learning-based methods extend geometry-driven navigation by introducing semantic perception, enabling target-aware exploration. For ground robots, modular approaches enrich classical pipelines with learned semantic sensing for object-goal navigation~\cite{chaplot2020objectgoal, das2024motion, gervet2023navigating}. Similarly, UIVNav~\cite{lin2024uivnav} applies imitation learning to underwater navigation, steering toward objects of interest. However, these methods depend on training data and may degrade in unseen environments or under appearance shifts such as turbid underwater conditions. Moreover, reactive learned policies lack long-horizon planning and high-level semantic reasoning needed to systematically explore complex environments where the goal is comprehensive coverage of distributed targets rather than single object instances.

\subsection{Vision-Language Models for Underwater Navigation}
\label{sec:rw_vlm}

Vision-language models (VLMs) offer open-vocabulary semantic understanding and stronger reasoning than learning-based methods, making them well-suited for interpreting complex scene semantics and making high-level decisions in unseen scenarios~\cite{firoozi2023foundation_robotics}. This is particularly valuable for ecological monitoring, where targets like "oyster reef clusters" cannot be pre-defined. While VLMs have been explored for ground robots, their underwater application remains limited. OceanChat~\cite{yang2023oceanchat} and OceanPlan~\cite{yang2024oceanplan} use text-only LLMs for AUV command interpretation but lack visual perception and adaptive exploration. However, VLMs are not ideal as standalone navigation controllers~\cite{kambhampati2024llmmodulo}. Their inference is slow, they do not model robot dynamics, and their outputs can be spatially coarse or unsafe~\cite{chen2024spatialvlm, liu2025spatial_survey}. VLMs also suffer from hallucination~\cite{liu2024hallucination_survey} and limited self-correction~\cite{huang2024selfcorrect}, causing errors to accumulate. DREAM~\cite{wu2025dream} addresses reef monitoring with VLM-guided exploration and Chain-of-Thought reasoning~\cite{wei2022cot}, achieving significant improvements over traditional baselines. However, DREAM queries the VLM at every step, coupling high-level planning with low-level execution. This causes frequent collisions due to slow VLM inference and accumulating path errors when the VLM fails to recover from obstacle-avoidance deviations.
To bridge these gaps, \sys~uses the VLM for semantic planning and target selection, coupled with DDP~\cite{lu2025ddp} for real time local planning, obstacle avoidance, and executable motion, with a geometric verification layer for self-correction.


%% file: sections/approach.tex
\section{Approach}
\label{sec:approach}
\begin{figure*}[t]
    \centering
    \includegraphics[width=0.99\textwidth]{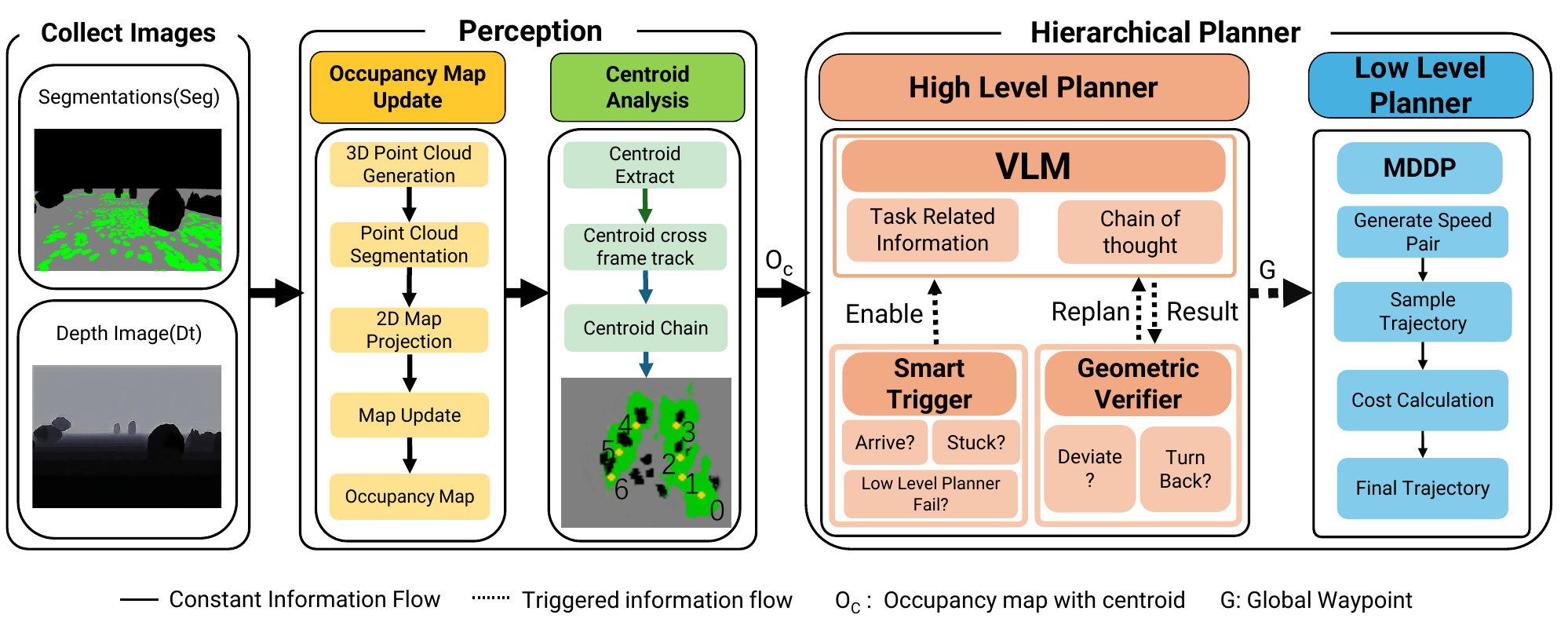}
    \caption{Overview of the \sys~framework. The perception module fuses segmentation and depth images into an occupancy map with labeled cluster centroids. The high-level planner invokes the VLM only when the smart trigger fires; a self-verification module rejects invalid waypoints with corrective feedback. Verified waypoints are executed by the MDDP low-level controller.}
    \label{fig:overview}
    \vspace{-5mm}
\end{figure*}
As depicted in Fig.~\ref{fig:overview}, \sys~consists of two major component of perception and planner. A perception module first fuses sensor data into a persistent occupancy map, from which a centroid extraction module identifies oyster clusters and organizes them into an ordered chain, transforming continuous exploration into discrete selection. The VLM high-level planner then reasons over this chain to choose the next target, whose validity is checked by formal geometric verifiers before it reaches the robot. To avoid unnecessary AI queries, a smart trigger ensures the VLM is consulted only when the robot reaches its goal, gets stuck, or encounters planner failure. Meanwhile, the MDDP controller runs continuously, executing collision-free trajectories at all times; the VLM periodically supplies it with updated global waypoints through these trigger events.
\vspace{-1mm}

\subsection{Perception and Occupancy Map}
\label{sec:perception}

The occupancy map is the shared spatial memory that all downstream modules, centroid extraction, VLM planning, and low-level control operate on. At each time step, the ROV's front-facing camera provides depth and semantic segmentation images. In simulation, ground-truth segmentation masks and depth maps are rendered directly from Blender, providing noise-free perception. In real-world trials, semantic segmentation is performed by YOLO26-seg and depth is estimated by Depth Anything V2~\cite{depthanythingv2}. Then, we incrementally project depth pixels into 3D world coordinates, classify them as targets (oysters), obstacles, or free space using the segmentation mask, and accumulate these observations into a persistent 2D grid map. The map is color-coded for both the VLM and human interpretability: green for detected oyster targets, black for obstacles, gray for explored free space, and white for unexplored regions.

Given a pixel $(u,v)$ with depth $z(u,v)$ and intrinsics $(f_x,f_y,c_x,c_y)$, its 3D position in the world frame is obtained by
\begin{equation}
\mathbf{p}_w(u,v) \;=\; 
\Pi\!\left(
\mathbf{T}_{c}^{w}
\begin{bmatrix}
\frac{(u-c_x)}{f_x}\,z(u,v) \\[6pt]
\frac{(v-c_y)}{f_y}\,z(u,v) \\[6pt]
z(u,v) \\[6pt]
1
\end{bmatrix}
\right),
\label{eq:backproj_world}
\end{equation}
where $\Pi$ denotes the homogeneous-to-Cartesian projection and $\mathbf{T}^w_c \in SE(3)$ is the camera-to-world transform. The resulting 3D points are partitioned into target objects $\mathcal{P}_{\text{obj}}$ (where the semantic mask $m(u,v)=1$) and obstacles $\mathcal{P}_{\text{obs}}$ (where $m(u,v) \neq 1$). To track which areas the robot has observed, a ray-casting process incrementally updates the explored mask by casting $N$ rays from the camera position along the robot heading within the horizontal field of view up to a maximum sensing range $d_{\max}$, unioning newly visible areas into the cumulative explored region.
\subsection{Centroid Chain Extraction and Tracking}
\label{sec:centroid}

Marine ecology shows that oysters form distinct clusters on reef substrates rather than being uniformly distributed, mainly because larvae settle on existing shell material~\cite{grabowski2012economic, cbf_oyster}. We use this prior to constrain the VLM action space: instead of predicting arbitrary coordinates, the VLM selects from extracted cluster centroids, ensuring each valid choice corresponds to a real oyster cluster. This reduces error while preserving autonomous target selection based on exploration progress and reef topology. The design is further motivated by evidence that VLMs are weak at quantitative spatial reasoning but more reliable at categorical selection with labeled visual options~\cite{chen2024spatialvlm}. We therefore extract cluster centroids from the perception output, organize them into an ordered chain, and present them as labeled candidates for VLM selection

\textbf{Centroid computation.}
For each oyster cluster identified on the map, let $r = \{(x_i, y_i)\}$ denote the set of pixels belonging to that cluster. Its centroid is computed as the spatial mean of all member pixels:
\begin{equation}
    \mathbf{c}_r = \frac{1}{|r|} \sum_{(x,y) \in r} (x, y),
\end{equation}
where $|r|$ is the number of pixels in the cluster, yielding one representative point per cluster.

\textbf{Connected component analysis.} We apply connected component labeling to the green mask $M_{\text{green}}$ to identify oyster clusters and compute their region properties. Regions with area below $A_{\min}$ are removed as noise, and the centroid of each remaining component is retained, producing an unordered set of cluster centroids.

\textbf{Path ordering.} We then organize these centroids into an ordered chain using a nearest neighbor greedy algorithm with union find path management. Starting from an arbitrary centroid, the algorithm repeatedly connects each path endpoint to its closest unvisited centroid and merges path segments, yielding an ordered sequence $\mathcal{C} = [c_0, c_1, \ldots, c_{K-1}]$ that approximates the spatial spine of the oyster distribution.

\textbf{Cross-frame tracking.}
As the robot moves and the occupancy map expands, new green regions appear and existing ones grow, causing raw centroid positions to shift between frames. Without tracking, the VLM would see centroid indices reassigned every step, making consistent planning impossible. We maintain a \texttt{CentroidTracker} that associates newly extracted centroids with existing tracks across frames. At each update, the tracker computes pairwise distances between all new centroids and all active tracks. If the nearest active track is within a merge distance $d_{\text{merge}}$, the new centroid is matched to that track and its position is smoothed via exponential moving average:
\begin{equation}
    \mathbf{c}_{\text{track}}^{t} = \alpha_{\text{old}} \cdot \mathbf{c}_{\text{track}}^{t-1} + \alpha_{\text{new}} \cdot \mathbf{c}_{\text{new}},
\end{equation}
where $\alpha_{\text{old}}$ and $\alpha_{\text{new}}$ are smoothing weights ($\alpha_{\text{old}} + \alpha_{\text{new}} = 1$). If no existing track is within $d_{\text{merge}}$, a new track is initialized. Tracks that are not matched for a configurable number of consecutive frames are removed. This ensures that (1)~centroid indices remain stable across frames so the VLM can reason consistently, (2) positions are gradually refined as more of the
cluster becomes visible, and (3) transient noise from partial
occlusion or incremental map updates does not create spurious
centroids. With a stable set of tracked centroids in hand, we convert them into a format suitable for both visual presentation to the VLM and metric navigation by the low-level controller.

\subsection{VLM-Based High-Level Planning}
\label{sec:vlm_planner}

Because a single centroid cannot adequately represent elongated or irregularly shaped clusters, the high-level planner operates in two complementary modes that switch autonomously based on cluster geometry: centroid selection for compact clusters and free waypoint prediction for elongated ones. This dual-mode design addresses a fundamental limitation of centroid-only guidance

\subsubsection{Centroid Selection Mode}
When multiple oyster clusters are present, the planner receives the occupancy map with labeled centroids and robot pose and trajectory overlay, the front camera segmentation image, and a structured prompt. The VLM returns a JSON dictionary with a selected centroid index and reasoning. The index is then converted to its odometry frame coordinate and sent to the DDP MPPI controller as the global waypoint.

The prompt is designed to improve reliability by framing the task as centroid index selection, describing the sensor inputs and their conventions, guiding step by step reasoning over segmentation, depth, robot orientation, trajectory history, and unexplored regions, and enforcing three rules: avoid areas near the yellow trajectory, prefer centroids near white unexplored regions, and follow the centroid chain direction. By restricting the output to a valid centroid index, the VLM avoids reasoning over continuous coordinates.

\subsubsection{Free Waypoint Mode}
When oyster clusters form elongated, strip-like structures (e.g., a continuous reef ridge), the centroid chain degenerates, centroids collapse to the interior of the strip, providing no directional guidance beyond the cluster center. In this scenario, the planner autonomously switches to a \emph{free waypoint mode} with a different prompt structure.

In this mode, the VLM receives the same sensor inputs (occupancy map with the centroid chain rendered as orange dots connected by blue lines, segmentation, and depth images) but is tasked with outputting next waypoint in the robot frame: a forward/backward distance and a left/right distance, each selected from discrete options $\{0, 1, 2, 3, 4, 5, 6\}$\,m rather than continuous coordinates, simplifying output parsing and constraining the error space.  The prompt instructs the VLM to: (a)~analyze the segmentation image for oyster distribution patterns; (b)~use the orange centroid chain on the occupancy map as a directional guide, planning waypoints \emph{along} the chain's trend; (c)~use depth information to calibrate distances; and (d)~verify the plan against the global occupancy map.

\subsubsection{Autonomous Mode Switching}
The system monitors the centroid chain geometry to determine the active mode. When no centroid exists in front of the robot, the planner switches from centroid selection to free waypoint mode. Conversely, when the robot discovers new clusters, detected by the appearance of multiple distinct connected components in the green mask, the planner switches back to centroid selection mode. This dual-mode design ensures efficient navigation both \emph{between} clusters (centroid mode) and \emph{along} extended reef structures (free waypoint mode).

\subsection{Formal Output Verification}
\label{sec:verification}

Even with structured prompts, VLMs produce invalid waypoints in a non-negligible fraction of queries. We implement two complementary verifiers that operate on the VLM's selected waypoint $\mathbf{p}_{\text{new}}$ in odometry coordinates. Let $\mathbf{p}_r$, $\theta_r$ denote the robot's current position and heading, $\vec{f}(\theta_r)$ the forward unit vector, and $\mathbf{p}_{\text{prev}}$ the previous target waypoint.

\subsubsection{Backward Detection}
This verifier rejects waypoints that would direct the robot backward. We define:
\begin{equation}
    \text{BehindRobot}(\mathbf{p}) \equiv \angle\big(\overrightarrow{\mathbf{p}_r \mathbf{p}},\; \vec{f}(\theta_r)\big) > \theta_{\text{back}},
    \label{eq:behind_robot}
\end{equation}
where $\theta_{\text{back}}$ is the backward rejection threshold. This check is performed at both the planning request time and the VLM response time, to account for robot motion during inference latency. Additionally, we check whether $\mathbf{p}_{\text{new}}$ falls behind the previous target:
\begin{equation}
    \text{BehindPrev} \equiv \neg\text{DistExempt} \;\wedge\; \angle(\vec{n}, \vec{b}) < \theta_{\text{behind}},
    \label{eq:behind_prev}
\end{equation}
where $\vec{b} = \text{norm}(\mathbf{p}_r - \mathbf{p}_{\text{prev}})$, $\vec{n} = \text{norm}(\mathbf{p}_{\text{new}} - \mathbf{p}_{\text{prev}})$, $\theta_{\text{behind}}$ is the behind-previous threshold, and the distance exemption $d_{\text{exempt}}$ permits far-away points. An exemption is also granted when no forward centroids exist, allowing turnaround at dead ends.

The combined backward rejection is:
\begin{equation}
    \text{RejectBackward} \equiv \text{BehindRobot} \;\lor\; \text{BehindPrev}.
\end{equation}

\subsubsection{Deviation Detection}
This verifier ensures the selected waypoint follows the centroid chain's spatial trend. To establish a baseline direction of progress along the centroid chain, we select
the farthest forward centroid $p_{\text{fwd}}$ (with forward projection $\geq d_{\text{fwd}}$ and lateral offset $< d_{\text{lat}}$) as a reference. The reference direction is defined as:
\begin{equation}
    \vec{v}_{\text{ref}} = \text{norm}(\mathbf{p}_{\text{fwd}} - \mathbf{p}_r),
\end{equation}
pointing from the robot toward the forward reference centroid (or from the backward reference toward the robot if no forward reference exists). The deviation predicate is:
\begin{equation}
    \text{IsDeviated} \equiv \angle(\vec{v}_{\text{ref}},\; \vec{v}_{\text{wp}}) > \theta_{\text{dev}},
    \label{eq:deviation}
\end{equation}
where $\vec{v}_{\text{wp}} = \text{norm}(\mathbf{p}_{\text{new}} - \mathbf{p}_r)$ and $\theta_{\text{dev}}$ is the maximum allowable deviation angle. This check is activated after sufficient map information has been accumulated and is skipped if the waypoint coincides with the reference centroid.

\subsubsection{Rejection and Feedback Loop}
When either verifier triggers, the waypoint is discarded and a structured rejection message is returned to the VLM, specifying the failure mode (e.g., ``selected centroid is behind the robot'' or ``selected centroid deviates to the left of the chain trend''). The VLM is then re-queried with this feedback appended, forming a generate-verify-correct loop that leverages external geometric reasoning rather than relying on the VLM's intrinsic self-correction ability.

\subsection{MDDP Low-Level Controller}
\label{sec:ddp}

Once a verified waypoint $\mathbf{p}^*_{\text{target}}$ is established, the low-level controller must generate a collision-free, dynamically feasible trajectory that respects the BlueROV2's hydrodynamic constraints. We adopt MDDP~\cite{lu2025ddp}, which addresses the classical gap between dynamics-free global planning and dynamics-aware local planning.

Conventional navigation systems use static dynamics parameters $\theta$---either full dynamics (local planner) or none (global planner). DDP generalizes this to time-varying parameters $\theta_t$:
\begin{equation}
    s_{t+1} = f(s_t, u_t;\; \theta_t),
    \label{eq:ddp_dynamics}
\end{equation}
where the dynamics integration interval increases along the rollout:
\begin{equation}
    \Delta_t = T \cdot \left[\left(\frac{t+1}{\mathcal{T}}\right)^p - \left(\frac{t}{\mathcal{T}}\right)^p\right],
    \label{eq:ddp_interval}
\end{equation}
and the number of collision-checking boundary points decreases as:
\begin{equation}
    N_t = n \cdot \left(1 - \left(\frac{t}{\mathcal{T}}\right)^p\right),
    \label{eq:ddp_points}
\end{equation}
where $T$ is the rollout time, $\mathcal{T}$ is the number of time steps, $n$ is the full set of boundary points, and $p$ is a hyperparameter controlling the decrement rate.

This design preserves high fidelity dynamics and dense collision checking near the robot, where reef navigation requires precise maneuvering, while simplifying the model farther along the trajectory to extend the planning horizon toward the VLM selected goal. The MPPI controller samples control sequences, rolls out trajectories with DDP's variable fidelity dynamics model, evaluates them using costs for goal progress, obstacle proximity, and smoothness, and computes the final control by weighted averaging over the lowest cost collision free trajectories. As a result, the system provides robust obstacle avoidance and ensures that planned trajectories remain physically executable for the underwater robot.

\subsection{Smart Trigger Mechanism}
\label{sec:smart_trigger}

In DREAM, the VLM is queried at every step, creating computational overhead and risking target oscillation when consecutive responses disagree. We replace this continuous invocation with condition-based triggering: the DDP-MPPI controller runs at all times, and the VLM is consulted only when one of the following conditions is met:
\begin{equation}
    s_{\text{plan}} = \text{Dist} \;\lor\; \text{Stuck} \;\lor\; \text{Fail},
    \label{eq:trigger}
\end{equation}
where the three trigger predicates are:

\noindent\textbf{Distance trigger:} $d(\mathbf{p}_{\text{robot}}, \mathbf{p}_{\text{target}}) < D_{\text{trig}}$, where $\mathbf{p}_{\text{robot}}$ is the robot's current position, $\mathbf{p}_{\text{target}}$ is the active goal waypoint, and $D_{\text{trig}}$ is the proximity threshold --- the robot has nearly reached its current goal and requires a new waypoint.

\noindent\textbf{Stuck trigger:} $\|\Delta \mathbf{p}\| < \Delta p_{\text{stuck}}$ and $|v| < v_{\text{stuck}}$ persisting for $> T_{\text{stuck}}$, where $\Delta \mathbf{p}$ is the position change over a monitoring window, $v$ is the current velocity magnitude, and $\Delta p_{\text{stuck}}$, $v_{\text{stuck}}$, $T_{\text{stuck}}$ are the respective thresholds --- the robot is unable to make progress toward its goal.

\noindent\textbf{Failure trigger:} $N_{\text{fail}} \geq N_{\text{fail\_max}}$, indicating that the current waypoint may be unreachable after too many consecutive planner failures. To avoid oscillation between waypoints, a cooldown period $T_{\text{cooldown}}$ sets a minimum interval between VLM queries, and a distance threshold $\Delta d_{\min}$ requires each new waypoint to provide meaningful spatial progress. The stuck trigger bypasses the cooldown to allow prompt recovery from trapped states.

%% file: sections/experiments.tex
\section{Experimental Setup}
\label{sec:experiments}

\subsection{Simulation Environment}

Based on open-source Oystersim~\cite{lin2022oystersim} and DREAM environments~\cite{wu2025dream}, we design 10 new Blender environments with progressively increasing complexity across five reef topology classes: L- and S-shaped layouts for basic and continuous turning, K-shaped intersections for branch selection, E-shaped structures with multiple prongs and dead ends for broader coverage, and O-shaped closed loops for continuous path following and return behavior.

The oyster densities are randomized within each layout to test the robustness of our model. Obstacle placement is also randomized: denser obstacles are placed in the initial region to stress-test collision avoidance under cluttered conditions, while obstacles in later regions are interspersed among oyster clusters to simulate the natural co-occurrence of reef structures and debris found in real oyster habitats. Compared to DREAM~\cite{wu2025dream}, which used sparse obstacle layouts, our environments feature \textbf{significantly} higher obstacle density and much more constrained passages, providing a more rigorous test of collision avoidance capability. The dimensions of the environment range from $20\times20$\,m to $30\times30$\,m.

\subsection{Real-World Setup}

We deploy \sys{} on a BlueROV2 Heavy equipped with a front-facing camera, onboard-IMU, and Waterlinked Sonar3D-15 for obstacle avoidance. As dipiced in Fig.~\ref{fig:beauty_shot}, trials are conducted in a controlled pool environment (12\,ft diameter, 5\,ft depth) with oyster shells arranged in various patterns and floating balls placed to simulate obstacles. The perception and occupancy map modules run in real time on a PC with RTX5090 graphic card, while VLM queries are sent to cloud-based API endpoints.

\subsection{Baselines}

We compare \sys{} against two baselines. \textbf{DREAM}~\cite{wu2025dream} uses discrete VLM-directed movement actions at each step with a PD controller and no centroid chain, verification, or smart trigger. \textbf{\sys{} w/o Low-Level Planner} retains all high-level modules (centroid chain, verification, smart trigger) but replaces DDP-augmented MPPI with DREAM's PD controller, isolating the contribution of the low-level planner. \textbf{\sys{} (Full)} is the complete system.

\subsection{Ablations}

Starting from the full \sys{} system, we ablate individual components to isolate their roles. \textbf{w/o Self Verification} removes geometric verification, so VLM selected centroids are sent directly to the low level planner without checks or feedback driven requerying. \textbf{w/o Centroid Select} keeps the centroid chain visible, but has the VLM predict a waypoint from direction and distance instead of selecting a centroid index, testing discrete selection against free form coordinate generation. \textbf{Rule Based Selection} replaces the VLM planner with a nearest unvisited centroid heuristic while retaining the remaining modules, isolating the benefit of VLM based semantic reasoning over a geometry only policy.

\subsection{Metrics}

Because DREAM and \sys{} have fundamentally different execution models, we define mission time separately for each. DREAM operates synchronously following its original protocol~\cite{wu2025dream}: the robot executes a 1\,s movement action, then calls the VLM and waits before issuing the next action, giving $T^{\text{DREAM}} = N_{\text{steps}} \times (1 + \bar{t}_{\text{VLM}})$. \sys{} operates asynchronously: DDP-MPPI runs continuously at $\Delta t = 0.1$\,s control period while VLM inference occurs in the background, giving $T^{\text{\sys{}}} = N_{\text{steps}} \times \Delta t + T_{\text{idle}}$, where $T_{\text{idle}}$ is the cumulative idle time from consecutive verification rejections; in practice $T_{\text{idle}}$ is negligible as most waypoints pass verification on the first attempt. In all experiments, we use GPT 5.2 from OPENAI. The median latency per VLM call is 5\,s for DREAM and 10\,s for \sys{}.

\begin{figure*}[t]
    \centering
    \includegraphics[width=1.00\textwidth]{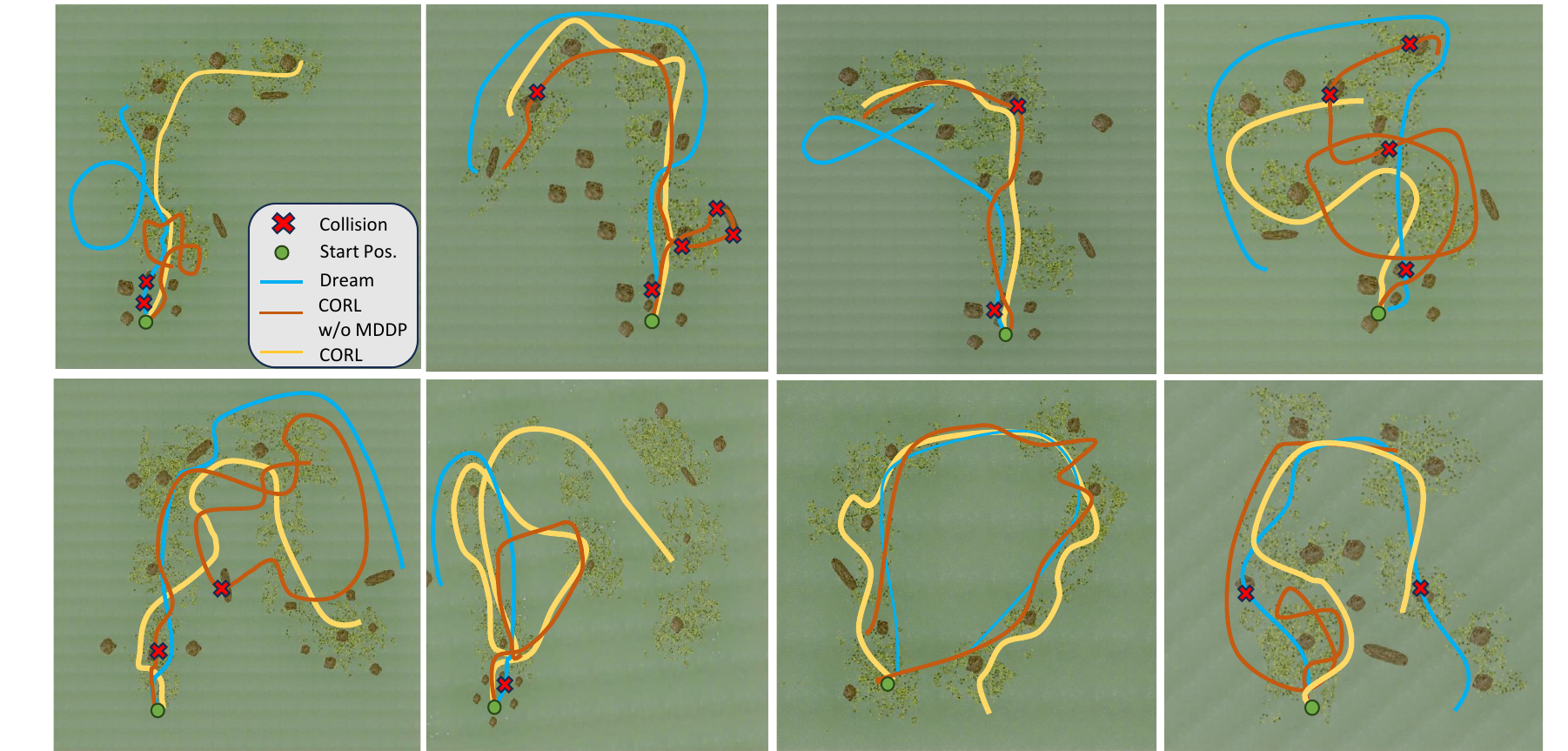}
    \caption{Trajectory comparison across 5 different topology representative environments. Blue: DREAM; orange: \sys{} w/o Low-Level Planner; yellow: \sys~(Full). DREAM's trajectories frequently deviate from oyster clusters and exhibit erratic motion due to per-step VLM control. \sys{} w/o Low-Level Planner follows the centroid chain more closely but produces jagged paths from PD control limitations. \sys~(Full)  achieves the smoothest, most coverage-efficient trajectories by combining VLM-guided waypoint selection with DDP-augmented MPPI execution.}
    \label{fig:trajectories}
    \vspace{-4mm}
\end{figure*}

We evaluate all methods on the following metrics. \textbf{Total coverage rate} (\%): fraction of ground-truth oyster clusters with at least 60\% of their area observed within the $120^\circ$ camera FOV, $N_{\text{cov}}/N_{\text{total}} \times 100\%$; higher is better. \textbf{Coverage time} (s): mission time to cover $K_i$ clusters in environment $i$, where $K_i$ is DREAM's final cluster count in that environment; lower is better. \textbf{Coverage efficiency} (steps/1\%): the number of steps required for every 1\% increase in coverage, $\eta = S_{\text{total}}/R_{\text{cov}}$, where $S_{\text{total}}$ is the total steps executed and $R_{\text{cov}}$ is the final coverage rate; a lower value indicates more efficient exploration. \textbf{Collision count}: total obstacle collisions; lower is better. \textbf{VLM invocations}: total VLM calls per mission; lower is better. \textbf{Deviation count}: the number of times the robot's trajectory deviates from the intended exploration route, registered when a waypoint violates the backward or off-trend constraints defined in Section~\ref{sec:verification}; lower is better.

\subsection{Implementation Details}
\label{sec:impl}

All experiments share the following configuration. The main pipeline runs at 10\,Hz ($\Delta t = 0.1$\,s). The smart trigger fires under three conditions: (1) the robot comes within 2.5\,m of its current target; (2) the robot's position change falls below 0.1\,m and velocity below 0.1\,m/s for more than 20\,s; or (3) the low-level planner fails twice consecutively. A cooldown of 50\,s and a minimum target separation of 1.0\,m prevent rapid waypoint switching. Centroid extraction filters connected components below 100\,px.

The DDP-MPPI controller samples 400 rollouts of 10 steps each, with linear and angular velocity perturbation standard deviations of 0.05. The robot radius is 0.3\,m and the obstacle sensing range is 4.0\,m. Cost weights are: goal 0.6, obstacle 0.5, and speed 0.3.

%% file: sections/results.tex
\section{Results}
\label{sec:results}
\vspace{8pt}

\begin{table}[t]
\centering
\caption{Baseline comparison across 10 simulation environments.  $\downarrow$: lower is better; $\uparrow$: higher is better.}
\label{tab:baseline}
\renewcommand{\arraystretch}{1.15}
\begin{tabular}{@{}l c c c c@{}}
\toprule
\textbf{Method} & \textbf{Cov.}$\uparrow$ & \textbf{Cov -Time}$\downarrow$ & \textbf{Coll.}$\downarrow$ & \textbf{VLM\_calls}$\downarrow$ \\
\midrule
DREAM                          & 80.00\% & 1513.2\,s  & 9    & 1261 \\
Ours w/o Low-Level    & 91.40\% & 687\,s  & 24 & 770 \\
\textbf{Ours (Full)}              & 94.28\% & 642\,s            & 0    & 547  \\
\bottomrule
\vspace{-8mm}
\end{tabular}
\end{table}    

\subsection{Baseline Comparison}

Table~\ref{tab:baseline} compares three configurations across all 10 simulation environments. DREAM, which couples VLM inference with every step, achieves 80.00\% coverage with 9 collisions and 1261 VLM calls. Adding \sys's high-level modules (centroid chain, verification, smart trigger) while retaining the original PD controller raises coverage to 91.4\% and reduces VLM calls to 770, confirming that the high-level architectural changes alone yield substantial gains. However, the PD controller incurs 24 collisions, nearly triple that of DREAM, because it lacks dynamics awareness and cannot plan collision-free trajectories through narrow passages. The full \sys{} system, replacing the PD controller with DDP-augmented MPPI, reaches 94.28\% coverage with zero collisions while requiring 57\% fewer VLM calls than DREAM.

As depicted in Fig.~\ref{fig:trajectories}, across eight representative test environments, the full method makes qualitatively more efficient exploration decisions than the baselines. From the common start position, its trajectory stays more tightly aligned with the reef structure, covers a larger portion of the traversable region, and reaches new areas with fewer unnecessary loops or sharp reversals. In contrast, Dream often makes large detours and drifts away from the informative regions, ~\sys{} tends to produce less stable paths with more local circling, and the variant without MDDP leaves coverage gaps or takes less direct routes to unexplored areas. The collision markers further show that the full method better balances safety and exploration, avoiding cluttered obstacles while still committing to wider, higher value branches of the environment. Overall, these examples suggest that the proposed design yields stronger long horizon reasoning and more effective reef coverage in diverse layouts.

\begin{table}[t]
\centering
\caption{Ablation study averaged across 10 environments. Each row modifies one component of the full system. Cov.: Coverage rate. Dev.: deviation count.}
\label{tab:ablation}
\renewcommand{\arraystretch}{1.15}
\begin{tabular}{@{}l c c c@{}}
\toprule
\textbf{Configuration} & \textbf{Cov.}$\uparrow$ & \textbf{Steps/1\%}$\downarrow$ & \textbf{Dev.}$\downarrow$ \\
\midrule
Ours (Full)                        & 94.28\% & 11.26  & 3 \\
\quad w/o Self Verification        & 67.14\% & 41.00  & 30\textsuperscript{$\dagger$} \\
\quad w/o Centroid Select          & 54.00\% & 23.63  & 15\textsuperscript{$\ddagger$} \\
Rule-Based Selection               & 51.06\% & 12.77  & $\infty$ \\
\bottomrule
\end{tabular}
\vspace{-6mm}
\end{table}

    \begin{figure}[t]
    \centering
    {\includegraphics[width=\columnwidth]{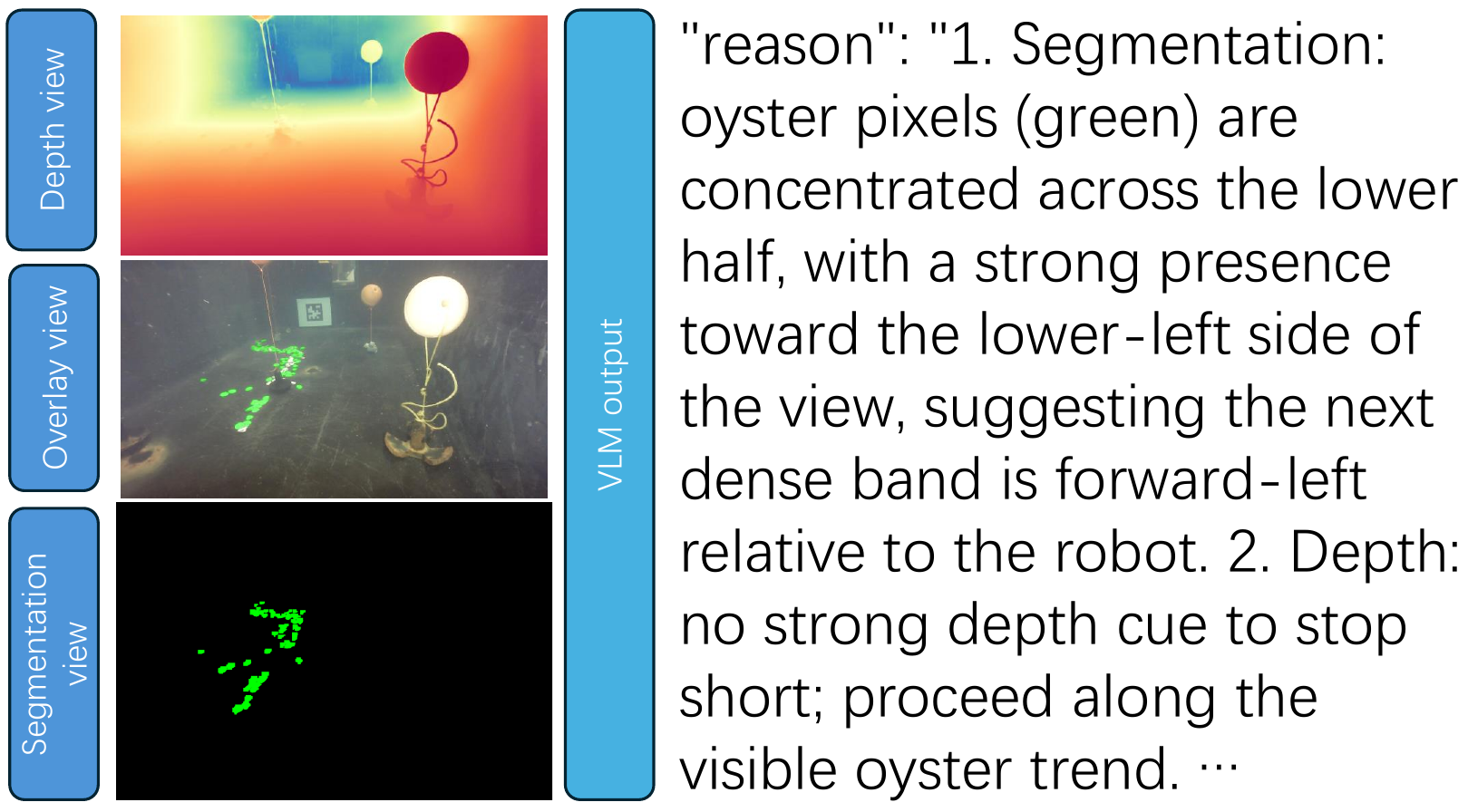}}
    \vspace{-3mm}
    \caption{Example of a real-world environment and VLM output }
    \label{fig:real-world-demo}
    \vspace{-4mm}
\end{figure}

\subsection{Ablation Study}

Table~\ref{tab:ablation} isolates each component's contribution. Removing self-verification nearly quadruples steps per unit coverage (11.26$\to$41) and causes persistent deviation in 3 environments, confirming that geometric checks are essential for route fidelity. Removing centroid selection forces the VLM to predict free-form coordinates, causing persistent deviation in 7 of 10 environments. The rule-based nearest-centroid heuristic achieves only 51.06\% coverage and exhibits highly inconsistent behavior across environments, indicating that geometric heuristics alone are insufficient for coherent reef exploration. We therefore focus our main comparisons on learning-based baselines that can leverage semantic cues and long-horizon decision making.

\subsection{Per-Topology Breakdown}

Table~\ref{tab:per_env} breaks down performance by reef topology. \sys~achieves 100\% coverage on L-shape, O-shape, and E-shape layouts, demonstrating robust handling of simple turns, closed loops, and multi-prong structures with dead ends. S-shape environments prove slightly more challenging at 92.86\%, as sustained double curves occasionally cause the VLM to miss a cluster near sharp transitions. K-shape topologies yield the lowest coverage (83.33\%) and demand the most VLM calls per environment (102.5), reflecting the difficulty of branch selection at intersections---the VLM sometimes commits to the wrong branch and must backtrack, consuming additional planning queries.

Notably, O-shape achieves the highest efficiency (1.26 steps/\%) despite requiring more VLM calls than simpler layouts, suggesting that closed-loop topologies naturally guide the robot along a productive path once the initial direction is established. E-shape environments, while reaching full coverage, exhibit the lowest efficiency (3.25 steps/\%) due to the cost of backtracking from dead ends to explore remaining prongs. Across all topologies, the DDP-MPPI controller maintains zero collisions, confirming that low-level obstacle avoidance is robust regardless of layout complexity.

\begin{table}[t]
\centering
\caption{Per-topology results for \sys.}
\label{tab:per_env}
\renewcommand{\arraystretch}{1.15}
\begin{tabular}{@{}l c c c c@{}}
\toprule
\textbf{Topology} & \textbf{Cov.}$\uparrow$ & \textbf{Steps/\%}$\downarrow$  & \textbf{VLM/Env}$\downarrow$ \\
\midrule
L-shape & 100\%   & 1.93   & 30 \\
S-shape & 92.86\% & 2.54   & 39 \\
O-shape & 100\%   & 1.26   & 65.5 \\
K-shape & 83.33\% & 2.17   & 102.5 \\
E-shape & 100\%   & 3.25   & 70 \\
\bottomrule
\end{tabular}
\end{table}   

\subsection{Failure Case Analysis}

Despite its strong overall performance, \textbf{CORAL} exhibits 
a recurring failure pattern observed across the 10 simulation 
environments.

\textbf{Global loop-back in branching topologies.} At branching 
reef structures where two plausible centroid chains diverge in 
different directions, the VLM occasionally selects a branch that 
passes both the backward-rejection and deviation checks locally, 
but globally leads back to previously explored regions. Because 
the geometric verifier evaluates only the immediate waypoint 
against local trajectory history, it cannot anticipate future 
path evolution. The VLM's single-step selection horizon is 
therefore insufficient to detect that a forward-facing branch 
will eventually converge back into the explored area, rendering 
the anti-backtracking mechanism ineffective in such cases. 
Addressing this limitation requires incorporating global 
topological path memory into the VLM context, enabling 
loop-closure awareness during waypoint selection.

\subsection{Real-world deployment result} 

A representative real-world trial is shown in 
Fig.~\ref{fig:real-world-demo}. The robot operates entirely 
without a prior map, incrementally building an occupancy 
representation from onboard segmentation and depth estimates 
as it moves. The VLM high-level planner selects successive 
centroid waypoints based on the evolving map, while the 
MDDP controller executes collision-free 
trajectories toward each goal. Throughout the trial, the robot 
successfully explore around all floating obstacles without 
contact, demonstrating that the low-level collision avoidance 
capbilities.

The robot achieved approximately 80--90\% oyster patch 
coverage over the course of the trial. Missed patches were 
primarily located behind obstacle anchor points, where 
physical occlusion and restricted viewing angles affect 
the detection, preventing centroid formation in those 
regions. 

%% file: sections/conclusion.tex
\section{Conclusion and Future Work}
\label{sec:conclusion}

We presented CORAL, a hierarchical framework that decouples VLM-based semantic exploration from dynamics-aware low-level control for autonomous underwater reef monitoring. By constraining high-level decisions to discrete centroid chains, correcting residual planning errors through geometric verification, and invoking the VLM only at meaningful decision points, CORAL enables reliable long-horizon exploration while maintaining continuous low-level control.

Across diverse reef topologies, CORAL achieves 94.28\% coverage with zero collisions, substantially outperforming DREAM, which achieves 80\% coverage with 9 collisions. Ablation studies further confirm that each design choice contributes to performance, highlighting the importance of structured high-level planning, self-verification, and efficient VLM invocation.

Future work will focus on extending CORAL to real-world open-water deployments with GPS-denied localization, replacing fixed-rule verification with learned adaptive verification models and prediction on future path. We also plan to explore multi-robot extensions for coordinated reef monitoring.

%% file: sections/appendix.tex
\newpage
\onecolumn
\appendix
\section{VLM Prompt Templates}
\label{sec:appendix}

The following prompts are used in all experiments. Prompt~A is used in free waypoint mode; Prompt~B is used in centroid selection mode.

\subsection{Prompt A: Free Waypoint Mode}

\begin{tcolorbox}[
  breakable,
  title={\textbf{Prompt A: High-Level Planner --- Next Waypoint Planning Guide}},
  fonttitle=\small\bfseries,
  colback=gray!5,
  colframe=black!60,
  coltitle=white,
  colbacktitle=black!70,
  left=4pt, right=4pt, top=4pt, bottom=4pt,
  fontupper=\small\ttfamily
]
\textbf{\#\# Role and Task}\\[4pt]
You are a \textbf{high-level planner}. Your task is to decide \textbf{where the robot should go next} by outputting a single \textbf{next waypoint} in the robot frame: forward distance and left/right distance in meters. A low-level controller executes the motion and handles obstacle avoidance; you only plan the next target point so the robot efficiently discovers and follows oyster clusters.\\[6pt]

\textbf{\#\# Sensor Input Description}\\[4pt]
\textbf{\#\#\# 1. Segmentation Image (Front Camera)}\\
- \textbf{Green}: Oysters (targets). \textbf{Gray}: Ground/seafloor.\\
- Analyze green pixel distribution: where oysters are, main direction (forward/left/right/diagonal), density, and whether they extend beyond the view.\\
- If green pixels form a line or cluster, plan the next waypoint along that pattern; if no green in a direction, no oysters there.\\[4pt]

\textbf{\#\#\# 2. Depth Image (Front Camera)}\\
- Distance from camera to scene (meters). Use depth at green (oyster) pixel locations to set \textbf{how far} the next waypoint should be (forward and left/right). Typical range 1--6 m. For diagonal plans, combine depth with segmentation for both components.\\[4pt]

\textbf{\#\#\# 3. Global Occupancy Map (Verification)}\\
- \textbf{Red dot}: Current robot position (origin of the robot frame on the map).\\
- \textbf{Green dots}: Historical oyster detections.\\
- \textbf{Orange dots + blue lines}: \textbf{Oyster centroid chain} --- each orange dot is the centroid of one green oyster cluster; blue lines connect them in spatial order. The chain is the \textbf{trend/spine} of oyster distribution; \textbf{prefer to plan waypoints along or extending this chain} when possible.\\
- \textbf{White}: Unexplored. \textbf{Gray}: Explored.\\
- \textbf{Verification}: Use the map to verify/correct the waypoint from segmentation (map is global, segmentation is local). In your reasoning you must describe the map in \textbf{three parts} --- see ``Thinking Process'' step 3.\\[6pt]

\textbf{\#\# Coordinate System and Output Range}\\[4pt]
Output is in the \textbf{robot frame} (origin = red dot on the map). \textbf{+X} = forward, \textbf{+Y} = left, \textbf{-X} = backward, \textbf{-Y} = right. One next waypoint: \textbf{Forward/Backward} and \textbf{Left/Right} in meters, each from discrete options \textbf{0, 1, 2, 3, 4, 5, 6} (typical range 1--6 m; use depth to choose). Always output both components; use \textbf{0} for a component if not needed (e.g. straight ahead $\rightarrow$ left/right = 0).\\[6pt]

\textbf{\#\# Strategy Summary}\\[4pt]
1. \textbf{Analyze} segmentation (oyster distribution), orange dots + blue lines (occupancy map), and depth (distance).\\
2. \textbf{Verify} with the occupancy map.\\
3. \textbf{Output} one next waypoint consistent with segmentation and map; \textbf{prefer to follow the orange centroid chain direction} when it indicates a clear trend.\\[6pt]

\textbf{\#\# Thinking Process}\\[4pt]
\textbf{1. Analyze the green (oyster) pixels in the segmentation image and the orange dots and blue lines (centroid chain) on the occupancy map}:\\
- \textbf{Segmentation}: Where are green (oyster) pixels? Initial distribution pattern: forward, left, right, diagonal, or behind? Density of oysters in different regions?\\
- \textbf{Orange centroid chain on occupancy map}: Where do the orange dots and blue lines run (sector, direction, left/right bias)? What trend do they indicate? \textbf{Goal}: Prefer to follow this chain direction when planning.\\
- Combine both: Preliminary next-waypoint direction --- \textbf{prioritize the orange centroid chain} as the main guide; use segmentation to refine.\\[4pt]

\textbf{2. Analyze Depth Image}:\\
- Depth at green (oyster) pixel locations.\\
- Average/median depth of oyster regions.\\[4pt]

\textbf{3. VERIFY with Occupancy Map (CRITICAL)}:\\
- \textbf{Describe the map in this order}:\\
\quad * \textbf{Point 1 (Orientation)}: Where do the red +X and blue +Y arrows point?\\
\quad * \textbf{Point 2 (Oyster distribution and centroid chain)}: Where do the green oyster dots lie and where does the \textbf{orange centroid chain} run (sector, direction, left/right bias, other clusters)?\\
\quad * \textbf{Point 3 (Unexplored regions)}: Where are contiguous white (unexplored) regions relative to the robot/oysters? Unfinished coverage?\\
- Check (a) green dots (oyster extent) and (b) orange centroid chain (cluster centers and trend).\\
- Compare map with your preliminary plan from step 1. Key questions: Does the full map confirm the direction from step 1? Is there a better trend elsewhere? If a turning trend is visible, plan the next waypoint to start the turn early.\\[4pt]

\textbf{4. Determine Final Next Waypoint Direction}:\\
- From BOTH segmentation and occupancy map, choose the direction for the next waypoint.\\
- \textbf{CRITICAL}: Plan waypoints \textbf{within or along the orange centroid chain} shown on the map. Do \textbf{not} plan waypoints far beyond the extent of the orange dots.\\
- \textbf{Priority}: Follow the \textbf{orange centroid chain} direction as much as possible. If the chain curves or turns, plan the next waypoint to follow that curve.\\
- \textbf{Avoid} waypoints that \textbf{severely deviate} from the centroid chain trend; stay aligned with the chain direction.\\
- Prefer straight ahead (+X) if oysters and orange centroid chain both indicate it. When segmentation and chain disagree, \textbf{favor the orange centroid chain} (global trend).\\[4pt]

\textbf{5. Set Waypoint Distance}:\\
- Use depth image to set forward (0--6 m, from the options).\\
- Set left/right distances (0--4  m, from the options).\\[4pt]

\textbf{6. Finalize}:\\
- Output the single next waypoint (forward, left/right) and state that you verified it using the occupancy map.\\[6pt]

\textbf{\#\# Output Format}\\[4pt]
Return ONLY a Python dictionary with exactly one key--value pair. The key is the \textbf{planned next waypoint} (action string), the value is your reason summary.\\[4pt]
\textbf{Example}:
\begin{verbatim}
{
  "Move forward 4 meters, move right 1 meter":
  "1. ...; 2. ...; 3. ...; 4. ...; 5. ...; 6. ...;
   Finally, I plan the next waypoint: forward 4 m, right 1 m."
}
\end{verbatim}
\end{tcolorbox}

\subsection{Prompt B: Centroid Selection Mode}

\begin{tcolorbox}[
  breakable,
  title={\textbf{Prompt B: High-Level Planner --- Select One Orange Dot as Next Waypoint}},
  fonttitle=\small\bfseries,
  colback=gray!5,
  colframe=black!60,
  coltitle=white,
  colbacktitle=black!70,
  left=4pt, right=4pt, top=4pt, bottom=4pt,
  fontupper=\small\ttfamily
]
\textbf{\#\# Role and Task}\\[4pt]
You are a \textbf{high-level planner}. Your task is to decide \textbf{where the robot should go next} by choosing \textbf{one orange dot} on the occupancy map as the next waypoint. Each orange dot is the centroid of an oyster cluster, labeled \textbf{0, 1, 2, \ldots} along the centroid chain. You must output the \textbf{index} of the orange dot you choose. A low-level controller will drive to that point; you only select which centroid to go to so the robot efficiently follows oyster clusters.\\[6pt]

\textbf{\#\# Sensor Input Description}\\[4pt]
\textbf{\#\#\# 1. Segmentation Image (Front Camera)}\\
- \textbf{Green}: Oysters (targets). \textbf{Gray}: Ground/seafloor.\\
- Analyze green pixel distribution: where oysters are, main direction (forward/forward left/forward right/left/right/diagonal), and density.\\
- \textbf{Oyster spatial trend (REQUIRED)}: Describe the trend of green oyster pixels across the image --- e.g., ``from bottom-left toward top-right'', ``from lower-center extending toward upper-left'', ``diagonal from left to right''. This trend should align your centroid choice with where oysters actually extend. Always state the trend explicitly.\\
- Use segmentation to help decide which orange dot aligns with visible oyster patterns.\\[4pt]

\textbf{\#\#\# 2. Depth Image (Front Camera)}\\
- Distance from camera to scene (meters). Use depth at green (oyster) pixel locations to understand how far oysters are.\\
- Prefer orange dots at reasonable distances based on depth information.\\[4pt]

\textbf{\#\#\# 3. Global Occupancy Map (Primary Input)}\\
- \textbf{Red dot + Red arrow (+X)}: Robot position and forward direction.\\
- \textbf{Blue arrow (+Y)}: Robot left direction. So \textbf{-Y} = robot right.\\
- \textbf{Yellow trajectory}: The path the robot has traveled --- this shows where you came from (the starting point of your trend).\\
- \textbf{Green dots}: Detected oyster detections.\\
- \textbf{Orange dots + blue lines}: Oyster centroid chain. Each orange dot is labeled \textbf{0, 1, 2, \ldots} in order along the chain. The chain shows the trend/spine of oyster distribution.\\
- \textbf{White}: Unexplored. \textbf{Gray}: Explored.\\[6pt]

\textbf{\#\# Strategy Summary}\\[4pt]
1. \textbf{Analyze} segmentation, depth, and the orange centroid chain on the occupancy map.\\
2. \textbf{Verify} with the occupancy map (orientation, distribution, unexplored regions).\\
3. \textbf{Choose} one orange dot that is close to the robot and near unexplored areas.\\[6pt]

\textbf{\#\# Thinking Process}\\[4pt]
\textbf{1. Segmentation Analysis}:\\
- Where are green (oyster) pixels? Distribution pattern: forward, left, right, diagonal?\\
- Density of oysters in different regions?\\[4pt]

\textbf{2. Depth Analysis}:\\
- Average depth of oyster regions --- helps decide if nearer or farther orange dots are preferred.\\[4pt]

\textbf{3. Occupancy Map Analysis (CRITICAL)}:\\
- \textbf{Point 1 (Orientation)}: Where do the red +X and blue +Y arrows point on the map?\\
- \textbf{Point 2 (Yellow trajectory)}: Where does the yellow trajectory go? This is where the robot came from --- the starting direction of the trend. Orange dots near or behind the yellow trajectory are ``already visited'' areas.\\
- \textbf{Point 3 (Oyster distribution and centroid chain)}: Where do green oyster dots lie? Where does the orange centroid chain (labels 0, 1, 2, \ldots) run? Use the yellow trajectory to determine the trend direction: the trend goes \textbf{away from} the yellow trajectory (from visited toward unvisited).\\
- \textbf{Point 4 (Unexplored regions)}: Where are white (unexplored) regions relative to robot and orange dots?\\[4pt]

\textbf{4. Choose Orange Dot}:\\
- \textbf{CRITICAL}: You \textbf{must} select an orange dot that \textbf{exists on the map}. The selected index must be within the range of labeled orange dots (0, 1, 2, \ldots). Do \textbf{not} output an index beyond the number of orange dots shown.\\
- \textbf{Priority rules} (in order):\\
\quad a. \textbf{Avoid} orange dots in areas the yellow trajectory has already passed through.\\
\quad b. Prefer orange dots \textbf{near unexplored (white) regions}.\\
\quad c. Prefer orange dots that \textbf{continue the trend away from} the yellow trajectory.\\
\quad d. Prefer orange dots that \textbf{follow the centroid chain direction}.\\
- The yellow trajectory shows where you came from; choose dots that move you toward new areas, not back to where you started.\\[4pt]

\textbf{5. Finalize}:\\
- Output \textbf{selected\_centroid\_index} and \textbf{reason} covering all analysis points above.\\[6pt]

\textbf{\#\# Output Format}\\[4pt]
Return ONLY a JSON-compatible Python dictionary with two keys:\\
- \textbf{selected\_centroid\_index} (int): 0-based index of the chosen orange dot.\\
- \textbf{reason} (string): Your step-by-step reasoning.\\[4pt]
\textbf{Example}:
\begin{verbatim}
{
  "selected_centroid_index": 2,
  "reason": "1. Segmentation: green oysters concentrated forward-right.
             2. Depth: oysters at ~x m. 3. Occupancy map: (1) Red +X
             points upper-left; (2) Yellow trajectory comes from
             lower-right, dots 0-1 are in visited area; (3) Orange chain
             runs toward upper-left, dots 2-5 are ahead; (4) White
             unexplored region ahead. 4. Selected orange dot 2."
}
\end{verbatim}
If there are \textbf{no orange dots} on the map:
\begin{verbatim}
{
  "selected_centroid_index": null,
  "reason": "No orange dots visible on the occupancy map."
}
\end{verbatim}
\end{tcolorbox}
The following are algorithms used in CORL, including main loop, high-level planner, low-level planner.
\begin{algorithm}[H]
\caption{CORAL Main Pipeline}
\label{alg:main}
\begin{algorithmic}[1]
\State \textbf{Input:} environment config, mission parameters
\State \textbf{Output:} trajectory, occupancy maps, mission logs
\Statex
\State \textcolor{gray}{$\triangleright$ \textit{Initialization}}
\State Initialize environment, occupancy map, navigation system, VLM planner
\State $\mathit{goal} \gets \text{initial waypoint}$;\enspace $\mathit{step} \gets 0$;\enspace reset trigger state
\Statex
\While{mission not complete}
    \Statex\hspace{\algorithmicindent}\textcolor{gray}{$\triangleright$ \textit{Perception}}
    \State Obtain depth and segmentation images from front camera
    \State Update occupancy map via ray-casting
    \Statex
    \hspace{\algorithmicindent}\textcolor{gray}{$\triangleright$ \textit{Smart Trigger}}
    \State $\mathit{trigger} \gets \mathit{DistTrigger} \lor \mathit{StuckTrigger} \lor \mathit{FailTrigger}$
    \If{$\mathit{trigger}$}
        \Statex
        \hspace{\algorithmicindent}\textcolor{gray}{$\triangleright$ \textit{VLM Planning (Alg.~\ref{alg:high})}}
        \State $\mathbf{p} \gets \text{VLMPlanner}(\text{map, seg, depth, robot pose})$
        \State Verify $\mathbf{p}$; re-query with feedback if rejected
        \State $\mathit{goal} \gets \mathbf{p}$
    \EndIf
    \Statex
    \hspace{\algorithmicindent}\textcolor{gray}{$\triangleright$ \textit{Low-Level Control (Alg.~\ref{alg:ddp})}}
    \State $(v^*, \omega^*) \gets \text{DDPMPPI}(\text{robot state}, \mathit{goal}, \text{map})$
    \State Execute $(v^*, \omega^*)$;\enspace $\mathit{step} \gets \mathit{step} + 1$
\EndWhile
\State \Return trajectory, maps, logs
\end{algorithmic}
\end{algorithm}

\begin{algorithm}[t]
\caption{VLM High-Level Planner}
\label{alg:high}
\begin{algorithmic}[1]
\Require occupancy map, segmentation image, depth image, robot pose
\Ensure verified waypoint $p^*$
\State $\triangleright$ \textit{Mode Selection}
\If{forward centroids exist beyond $d_{\min} = 1$ m}
    \State $\triangleright$ \textit{Centroid Selection Mode}
    \State $\text{idx} \leftarrow \text{VLM}(\text{map, seg, depth, prompt}_\text{centroid})$
    \State $p \leftarrow \text{centroid}[\text{idx}]$
\Else
    \State $\triangleright$ \textit{Free Waypoint Mode}
    \State $(d_\text{fwd}, d_\text{lat}) \leftarrow \text{VLM}(\text{map, seg, depth, prompt}_\text{free}),\ d \in \{0,\ldots,6\}$ m
    \State $p \leftarrow \text{RobotFrame}(d_\text{fwd}, d_\text{lat})$
\EndIf
\State $\triangleright$ \textit{Geometric Verification Loop}
\While{\text{RejectBackward}(p) \textbf{or} \text{IsDeviated}(p)}
    \State Append rejection feedback to prompt
    \State $p \leftarrow \text{VLM re-query with updated prompt}$
\EndWhile
\State \Return $p^* \leftarrow p$
\end{algorithmic}
\end{algorithm}

\begin{algorithm}[H]
\caption{MDDP Low-Level Controller}
\label{alg:ddp}
\begin{algorithmic}[1]
\State \textbf{Input:} robot state $(x, y, \theta, v, \omega)$, goal $\mathbf{p}^*$, occupancy map
\State \textbf{Output:} control action $(v^*, \omega^*)$
\Statex
\State \textcolor{gray}{$\triangleright$ \textit{Trajectory Sampling}}
\State Sample $N = 400$ control sequences $\{u_t\}_{t=0}^{\mathcal{T}-1}$
\For{each rollout $k = 1, \ldots, N$}
    \Statex\hspace{\algorithmicindent}\textcolor{gray}{$\triangleright$ \textit{Variable-Fidelity Rollout}}
    \For{$t = 0, \ldots, \mathcal{T}-1$}
        \State $s_{t+1} \gets f(s_t, u_t;\,\theta_t)$ with $\Delta_t \propto t^p$
        \State Check $N_t \propto (1-t/\mathcal{T})^p$ collision boundary points
    \EndFor
    \State $J_k \gets w_\text{goal}C_\text{goal} + w_\text{obs}C_\text{obs} + w_\text{speed}C_\text{speed} + w_\text{path}C_\text{path}$
\EndFor
\Statex
\State \textcolor{gray}{$\triangleright$ \textit{Control Output}}
\State $(v^*, \omega^*) \gets$ weighted average of lowest-cost feasible rollouts
\State \Return $(v^*, \omega^*)$
\end{algorithmic}
\end{algorithm}

%% file: references.bib
@inproceedings{lu2025ddp,
  title={Decremental dynamics planning for robot navigation},
  author={Lu, Yuanjie and Xu, Tong and Wang, Linji and Hawes, Nick and Xiao, Xuesu},
  booktitle={2025 IEEE/RSJ International Conference on Intelligent Robots and Systems (IROS)},
  pages={4559--4565},
  year={2025},
  organization={IEEE}
}

@article{xiao2022motion,
  title={Motion planning and control for mobile robot navigation using machine learning: a survey},
  author={Xiao, Xuesu and Liu, Bo and Warnell, Garrett and Stone, Peter},
  journal={Autonomous Robots},
  volume={46},
  number={5},
  pages={569--597},
  year={2022},
  publisher={Springer}
}

@inproceedings{lin2024uivnav,
  title={{UIVNav}: Underwater Information-Driven Vision-Based Navigation via Imitation Learning},
  author={Lin, Xiaomin and others},
  booktitle={Proc. IEEE Int. Conf. Robotics and Automation (ICRA)},
  pages={5250--5256},
  year={2024}
}

@inproceedings{lin2022oystersim,
  title={{OysterSim}: Underwater Simulation for Enhancing Oyster Reef Monitoring},
  author={Lin, Xiaomin and others},
  booktitle={OCEANS 2022, Hampton Roads},
  pages={1--6},
  year={2022},
  organization={IEEE}
}

@inproceedings{chen2024spatialvlm,
  title={{SpatialVLM}: Endowing Vision-Language Models with Spatial Reasoning Capabilities},
  author={Chen, Boyuan and Xu, Zhuo and Kirmani, Sean and Ichter, Brain and Sadigh, Dorsa and Guibas, Leonidas and Xia, Fei},
  booktitle={Proc. IEEE/CVF Conf. Computer Vision and Pattern Recognition (CVPR)},
  pages={14455--14465},
  year={2024}
}

@article{liu2025spatial_survey,
  title={Deconstructing Spatial Intelligence in Vision-Language Models: A Comprehensive Survey},
  author={Liu, Disheng and others},
  journal={TechRxiv preprint},
  year={2025}
}

@article{liu2024hallucination_survey,
  title={A Survey on Hallucination in Large Vision-Language Models},
  author={Liu, Hanchao and Xue, Wenyuan and Chen, Yifei and Chen, Dapeng and Zhao, Xiutian and Wang, Ke and Hou, Liping and Li, Rongjun and Peng, Wei},
  journal={arXiv preprint arXiv:2402.00253},
  year={2024}
}

@inproceedings{huang2024selfcorrect,
  title={Large Language Models Cannot Self-Correct Reasoning Yet},
  author={Huang, Jie and Chen, Xinyun and Mishra, Swaroop and Zheng, Huaixiu Steven and Yu, Adams Wei and Song, Xinying and Zhou, Denny},
  booktitle={Proc. Int. Conf. Learning Representations (ICLR)},
  year={2024}
}

@article{wei2022cot,
  title={Chain-of-Thought Prompting Elicits Reasoning in Large Language Models},
  author={Wei, Jason and Wang, Xuezhi and Schuurmans, Dale and Bosma, Maarten and Xia, Fei and Chi, Ed and Le, Quoc V and Zhou, Denny},
  journal={Advances in Neural Information Processing Systems},
  volume={35},
  pages={24824--24837},
  year={2022}
}

@article{yang2024oceanplan,
  title={{OceanPlan}: Hierarchical Planning and Replanning for Natural Language {AUV} Piloting},
  author={Yang, Ruochu and Zhang, Fumin and Hou, Mengxue},
  journal={Proc. Int. Conf. Underwater Networks \& Systems},
  year={2024}
}

@article{yang2023oceanchat,
  title={{OceanChat}: Piloting Autonomous Underwater Vehicles in Natural Language},
  author={Yang, Ruochu and Hou, Mengxue and Wang, Junkai and Zhang, Fumin},
  journal={arXiv preprint arXiv:2309.16052},
  year={2023}
}

@article{grabowski2012economic,
  title={Economic Valuation of Ecosystem Services Provided by Oyster Reefs},
  author={Grabowski, Jonathan H and Brumbaugh, Robert D and Conrad, Robert F and Keeler, Andrew G and Opaluch, James J and Peterson, Charles H and Piehler, Michael F and Powers, Sean P and Smyth, Ashley R},
  journal={BioScience},
  volume={62},
  number={10},
  pages={900--909},
  year={2012}
}

@misc{cbf_oyster,
  title={Oyster Reefs},
  author={{Chesapeake Bay Foundation}},
  howpublished={\url{https://www.cbf.org/nature/oyster-reefs/}},
  year={2025}
}

@article{beck2011oyster,
  author  = {Michael W. Beck and Robert D. Brumbaugh and Laura Airoldi and Alvar Carranza and Loren D. Coen and Christine Crawford and Omar Defeo and Graham J. Edgar and Boze Hancock and Matthew C. Kay and Hunter S. Lenihan and Mark W. Luckenbach and Caitlyn L. Toropova and Guofan Zhang and Ximing Guo},
  title   = {Oyster Reefs at Risk and Recommendations for Conservation, Restoration, and Management},
  journal = {BioScience},
  volume  = {61},
  number  = {2},
  pages   = {107--116},
  year    = {2011},
  doi     = {10.1525/bio.2011.61.2.5}
}

@article{howie2021contemporary,
  author  = {Alice H. Howie and Melanie J. Bishop},
  title   = {Contemporary Oyster Reef Restoration: Responding to a Changing World},
  journal = {Frontiers in Ecology and Evolution},
  volume  = {9},
  pages   = {689915},
  year    = {2021},
  doi     = {10.3389/fevo.2021.689915}
}

@article{zhang2023auvreview,
  author  = {Bingbing Zhang and Daxiong Ji and Shuo Liu and Xinke Zhu and Wen Xu},
  title   = {Autonomous Underwater Vehicle navigation: A review},
  journal = {Ocean Engineering},
  volume  = {273},
  pages   = {113861},
  year    = {2023},
  doi     = {10.1016/j.oceaneng.2023.113861}
}

@article{cardenas2024robotic,
  author  = {Jennifer A. Cardenas and Zahra Samadikhoshkho and Ateeq Ur Rehman and Alexander U. Valle-P{\'e}rez and Elena Herrera-Ponce de Le{\'o}n and Charlotte A. E. Hauser and Eric M. Feron and Rafiq Ahmad},
  title   = {A systematic review of robotic efficacy in coral reef monitoring techniques},
  journal = {Marine Pollution Bulletin},
  volume  = {202},
  pages   = {116273},
  year    = {2024},
  doi     = {10.1016/j.marpolbul.2024.116273}
}

@inproceedings{chaplot2020objectgoal,
  title={Object Goal Navigation Using Goal-Oriented Semantic Exploration},
  author={Chaplot, Devendra Singh and Gandhi, Dhiraj and Gupta, Abhinav and Salakhutdinov, Ruslan},
  booktitle={Advances in Neural Information Processing Systems (NeurIPS)},
  year={2020}
}

@article{christensen2022underwater,
  title={Recent Advances in {AI} for Navigation and Control of Underwater Robots},
  author={Christensen, Leif and de Gea Fern{\'a}ndez, Jos{\'e} and Hildebrandt, Marc and Koch, Christian Ernst Siegfried and Wehbe, Bilal},
  journal={Current Robotics Reports},
  volume={3},
  pages={165--175},
  year={2022}
}

@inproceedings{firoozi2023foundation_robotics,
  title={Foundation Models in Robotics: Applications, Challenges, and the Future},
  author={Firoozi, Roya and Tucker, Johnathan and Tian, Stephen and Majumdar, Anirudha and Sun, Jiankai and Liu, Weiyu and Zhu, Yuke and Song, Shuran and Kapoor, Ashish and Hausman, Karol and others},
  journal={The International Journal of Robotics Research},
  volume={44},
  number={5},
  pages={701--739},
  year={2023},
  publisher={SAGE}
}

@article{gervet2023navigating,
  title={Navigating to Objects in the Real World},
  author={Gervet, Th{\'e}ophile and Chintala, Soumith and Batra, Dhruv and Malik, Jitendra and Chaplot, Devendra Singh},
  journal={Science Robotics},
  volume={8},
  number={79},
  pages={eadf6991},
  year={2023},
  publisher={AAAS}
}

@inproceedings{kambhampati2024llmmodulo,
  title={Position: {LLMs} Can't Plan, But Can Help Planning in {LLM-Modulo} Frameworks},
  author={Kambhampati, Subbarao and Valmeekam, Karthik and Guan, Lin and Verma, Mudit and Stechly, Kaya and Bhambri, Siddhant and Saldyt, Lucas Paul and Murthy, Anil B},
  booktitle={Proc. Int. Conf. Machine Learning (ICML)},
  pages={22895--22907},
  year={2024}
}

@inproceedings{das2024motion,
  title={Motion memory: Leveraging past experiences to accelerate future motion planning},
  author={Das, Dibyendu and Lu, Yuanjie and Plaku, Erion and Xiao, Xuesu},
  booktitle={2024 IEEE International Conference on Robotics and Automation (ICRA)},
  pages={16467--16474},
  year={2024},
  organization={IEEE}
}

@article{lu2025adaptive,
  title={Adaptive Dynamics Planning for Robot Navigation},
  author={Lu, Yuanjie and Mao, Mingyang and Xu, Tong and Wang, Linji and Lin, Xiaomin and Xiao, Xuesu},
  journal={arXiv preprint arXiv:2510.05330},
  year={2025}
}

@article{xiaoautonomous,
  title={Autonomous Ground Navigation in Highly Constrained Spaces: Lessons Learned from The Forth BARN Challenge at ICRA 2025},
  author={Xiao, Xuesu and Xu, Zifan and Ghani, Saad Abdul and Datar, Aniket and Song, Daeun and Stone, Peter and Mazen, Amna and Yazdipaz, Kamyab and Mateyaunga, Innocent and Faied, Mariam and others}
}

@article{xu2025verti,
  title={Verti-bench: A general and scalable off-road mobility benchmark for vertically challenging terrain},
  author={Xu, Tong and Pan, Chenhui and Rao, Madhan B and Datar, Aniket and Pokhrel, Anuj and Lu, Yuanjie and Xiao, Xuesu},
  journal={arXiv preprint arXiv:2502.11426},
  year={2025}
}

@inproceedings{wu2025dream,
  title={{DREAM}: Domain-Aware Reasoning for Efficient Autonomous Underwater Monitoring},
  author={Wu, Zhenqi and Modi, Abhinav and Mavrogiannis, Angelos and Joshi, Kaustubh and Chopra, Nikhil and Aloimonos, Yiannis and Karapetyan, Nare and Rekleitis, Ioannis and Lin, Xiaomin},
  booktitle={Proc. IEEE Int. Conf. Robotics and Automation (ICRA)},
  year={2025}
}

@article{wynn2014auv_marine,
  title={Autonomous Underwater Vehicles ({AUVs}): Their Past, Present and Future Contributions to the Advancement of Marine Geoscience},
  author={Wynn, Russell B and Huvenne, Veerle AI and Le Bas, Timothy P and Murton, Bramley J and Connelly, Douglas P and Bett, Brian J and Ruhl, Henry A and Morris, Kirsty J and Peakall, Jeff and Parsons, Daniel R and others},
  journal={Marine Geology},
  volume={352},
  pages={451--468},
  year={2014},
  publisher={Elsevier}
}

@inproceedings{yamauchi1997frontier,
  title={A frontier-based approach for autonomous exploration},
  author={Yamauchi, Brian},
  booktitle={Proceedings 1997 IEEE International Symposium on Computational Intelligence in Robotics and Automation CIRA'97},
  pages={146--151},
  year={1997},
  organization={IEEE}
}

@article{depthanythingv2,
  title   = {Depth Anything V2},
  author  = {Yang, Lihe and Kang, Bingyi and Huang, Zilong and Zhao, Zhen and Xu, Xiaogang and Feng, Jiashi and Zhao, Hengshuang},
  journal = {arXiv preprint arXiv:2406.09414},
  year    = {2024}
}

@article{wang2025underwatervla,
  title={UnderwaterVLA: Dual-brain Vision-Language-Action architecture for Autonomous Underwater Navigation},
  author={Wang, Zhangyuan and Zhu, Yunpeng and Yan, Yuqi and Tian, Xiaoyuan and Shao, Xinhao and Li, Meixuan and Li, Weikun and Su, Guangsheng and Cui, Weicheng and Fan, Dixia},
  journal={arXiv preprint arXiv:2509.22441},
  year={2025}
}
